\documentclass[sigconf,authorversion,nonacm]{acmart}

\AtBeginDocument{%
  }

\setcopyright{acmlicensed}
\copyrightyear{2024}
\acmYear{2024}
\acmISBN{978-1-4503-XXXX-X/18/06}

\acmConference[CIKM '24]{33rd ACM International Conference on
Information and Knowledge Management}{October 21--25,
  2024}{Boise, ID}
\acmISBN{978-1-4503-XXXX-X/18/06}

\usepackage{algorithm}
\usepackage{algpseudocode}
\usepackage{float}
\usepackage{booktabs}
\usepackage{xcolor}
\begin{document}

\title{Context-Aware SQL Error Correction Using Few-Shot Learning - A Novel Approach Based on NLQ, Error, and SQL Similarity}

\author{Divyansh Jain}
\authornote{Both authors contributed equally to this research.}
\email{divyanshj@nvidia.com}
\orcid{0009-0006-1073-4608}
\author{Eric Yang}
\authornotemark[1]
\email{ericyang@nvidia.com}
\affiliation{%
  \institution{RAPIDS, NVIDIA}
  \city{Santa Clara}
  \state{California}
  \country{USA}
}


\begin{abstract}
In recent years, the demand for automated SQL generation has increased significantly, driven by the need for efficient data querying in various applications. However, generating accurate SQL queries remains a challenge due to the complexity and variability of natural language inputs. This paper introduces a novel few-shot learning-based approach for error correction in SQL generation, enhancing the accuracy of generated queries by selecting the most suitable few-shot error correction examples for a given natural language question (NLQ). In our experiments with the open-source Gretel dataset \cite{gretel-synthetic-text-to-sql-2024}, the proposed model offers a 39.2\% increase in fixing errors from the baseline approach with no error correction and a 10\% increase from a simple error correction method. 

The proposed technique leverages embedding-based similarity measures to identify the closest matches from a repository of few-shot examples. Each example comprises an incorrect SQL query, the resulting error, the correct SQL query, and detailed steps to transform the incorrect query into the correct one. By employing this method, the system can effectively guide the correction of errors in newly generated SQL queries.

Our approach demonstrates significant improvements in SQL generation accuracy by providing contextually relevant examples that facilitate error identification and correction. The experimental results highlight the effectiveness of embedding-based selection in enhancing the few-shot learning process, leading to more precise and reliable SQL query generation. This research contributes to the field of automated SQL generation by offering a robust framework for error correction, paving the way for more advanced and user-friendly database interaction tools.
\end{abstract}




\received{24 August 2024}
\received[revised]{15 October 2024}
\received[accepted]{4 September 2024}

\maketitle

\section{Related Work}
While there have been several approaches for error correction \cite{dong2023c3zeroshottexttosqlchatgpt, chen2023teachinglargelanguagemodels} , they rely on manually crafted guidelines by human experts, which are labor-intensive and limited by the human ability to identify all potential error patterns. On the other hand, while auto-correction approaches \cite{Askari2024MAGICGS} have shown promise, they often lack specific steps to correct errors or miss the context in which the NLQ was asked. Our work is an extension of the automatic SQL correction approach \cite{chen2023texttosqlerrorcorrectionlanguage} where predicted SQL are represented in token-level and clause-level format by using depth-first AST traversal. We take this a step further by representing the difference between all incorrect SQLs and their golden counterpart to create a few-shot example pool. Then we use embedding similarity to select the most-helpful few-shot correction examples for the target query and incorrect SQL using the technique outlined below.

Several studies have explored few-shot learning for text-to-SQL tasks \cite{yang2023harnessingpowerllmspractice, 10.1145/3544548.3581388}. For instance, SQLPrompt leverages execution-based consistency and diverse prompt designs to improve few-shot prompting capabilities \cite{10.1145/3589292}. Similarly, MCS-SQL employs multiple prompts and selection strategies to enhance the robustness of SQL generation \cite{sun2023sqlpromptincontexttexttosqlminimal}. These approaches highlight the importance of example selection and prompt design in improving SQL generation accuracy.

Empirical studies have demonstrated that few-shot prompting for text-to-SQL tasks, evaluated across multiple datasets and using various large language models (LLMs) \cite{pourreza2023dinsqldecomposedincontextlearning, zhang2024structureguidedlargelanguage}, shows solid performance compared to zero-shot prompting. For instance, providing a single detailed example (1-shot) \cite{li2023llmservedatabaseinterface} can effectively trigger a text-to-SQL model to generate accurate SQL queries. Additionally, research has examined the impact of the number of few-shot examples on performance. While this few-shot approach has shown promise in generating SQL, it remains largely untapped for SQL error correction. This presents an opportunity to explore its potential in improving the accuracy and reliability of SQL query generation by focusing on error correction strategies.

\section{Methodology}
\subsection{Initial Setup Process}
The initial setup involves creating a set of predefined examples that form the basis of our few-shot correction approach. For each NLQ in our dataset, we generate a predicted SQL query and compare it against the golden SQL. In cases where the predicted SQL is incorrect, we utilize the Change Distiller algorithm, developed by Fluri et al. \cite{4339230}, to compare the predicted and golden SQLs. Change Distiller analyzes the Abstract Syntax Trees (ASTs) of the SQL queries, identifying mismatched nodes and generating an edit script that details the required transformations to convert the incorrect SQL into the correct one. This process allows us to document not only the incorrect SQL but also the specific errors and the steps needed for correction.

As a result, our predefined examples include the NLQ, the incorrectly generated SQL, the correct (golden) SQL, any errors identified in the incorrect SQL, and the differences between the SQLs as determined by the Change Distiller edit script. These examples serve as the foundation for our few-shot selection process, ensuring that the chosen examples are contextually relevant and contain both the problem and the solution in a structured format. 

Figure 1 describes the creation of the few-shot Error Correction (EC) examples using the initial setup process.

% \onecolumn
\begin{figure}[h]
  \centering
  \includegraphics[width=\linewidth]{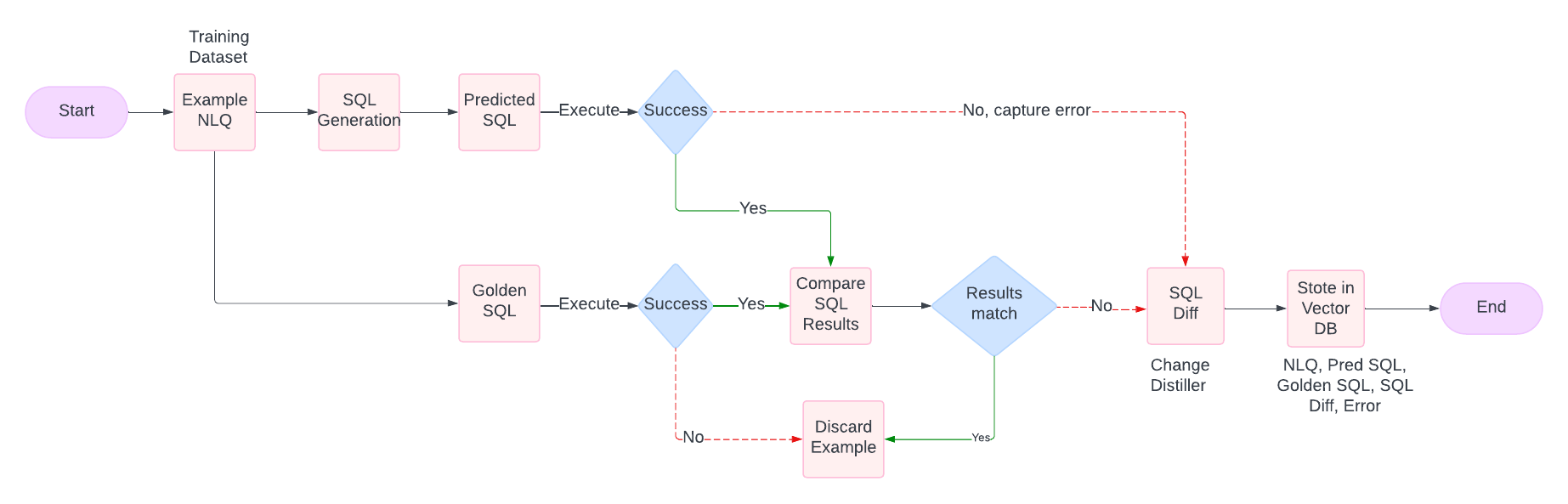}
  \caption{Initial setup process for few-shot Error correction. All examples except the discarded ones are saved to the Vector Store DB to be used later for few-shot selection. For saved examples, the Example NLQ, Predicted SQL, Golden SQL, Difference between the Predicted \& Golden SQL \& the Error (if captured) are stored as Embeddings.}
  \Description{A flow diagram for the setup process of the few-shot Error correction. The few-shot examples are stored in a vector database at the final step.}
\end{figure}

% \twocolumn
\subsection{Few-Shot Selection}
Once the predefined examples are established, our correction methodology proceeds by generating an SQL query based on a new NLQ and identifying any errors in the generated query. Embeddings for the NLQ, the generated incorrect SQL, and the resulting error are computed and compared against the set of predefined examples. The closest matching example is selected, which aligns with the error and includes the correct SQL and a detailed correction process. We tried different setups to identify the closest few-shot example which are explained in the experiment section.

\subsection{SQL Correction}
The selected example(s) are then provided as a few-shot example to correct the erroneous SQL along with the resulting error and the original NLQ. 

Figure 2 below is a representation of the Few-Shot Selection and SQL Correction process described in 2.2 and 2.3.

\begin{figure}[h]
  \centering
  \includegraphics[width=\linewidth]{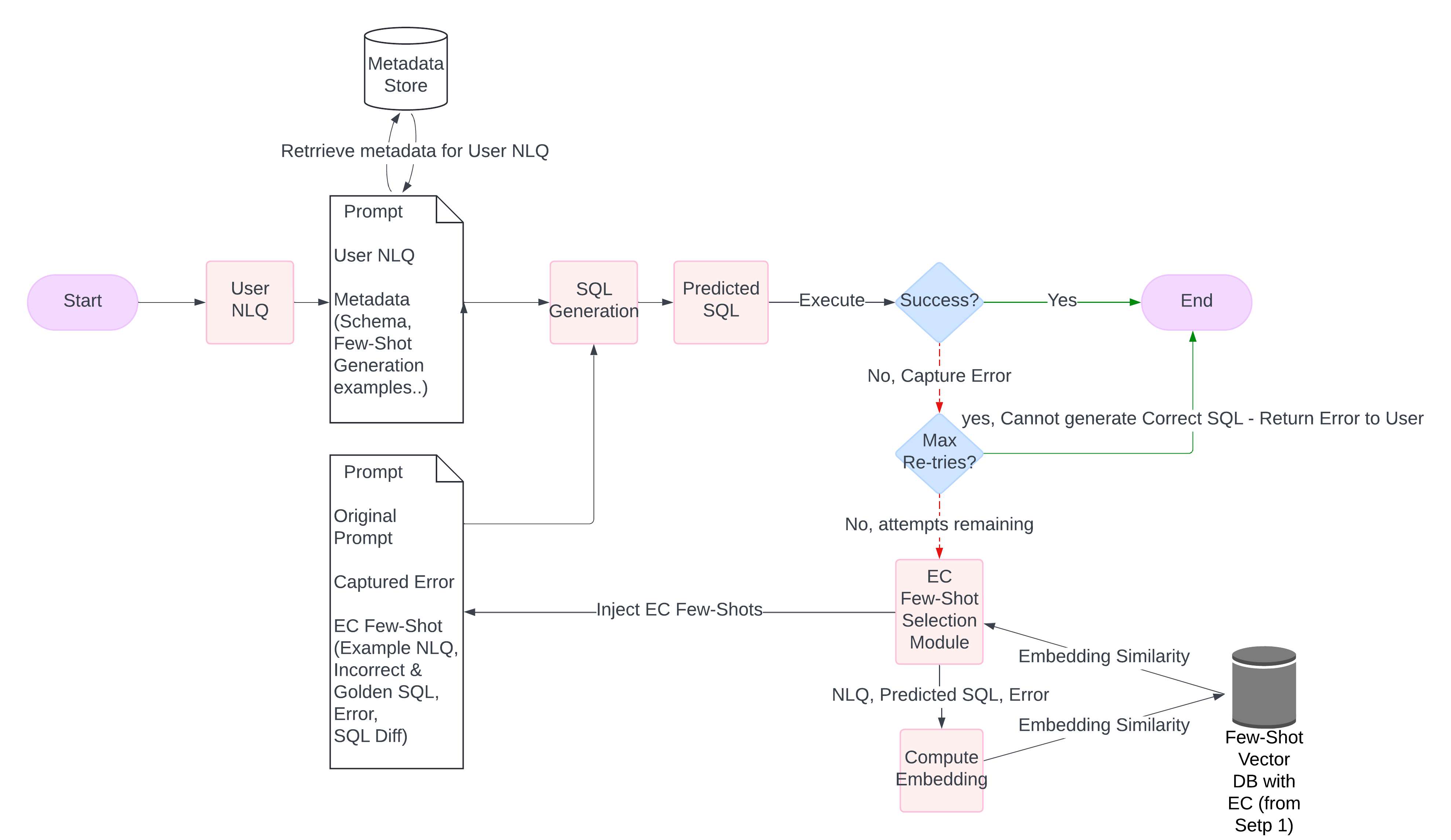}
  \caption{Few-Shot Selection \& SQL Correction using few-shot EC examples.}
  \Description{A flow diagram for the inference part of few-shot selection and SQL correction process.}
\end{figure}

\subsection{Proposed Algorithm}
We introduce two complementary algorithms designed to improve SQL error correction through a few-shot learning approach. Algorithm 1 focuses on generating a set of predefined examples by comparing predicted SQL queries with their correct counterparts, using the Change Distiller algorithm to create transformation scripts for incorrect predictions. 
\begin{algorithm}[H]
\caption{Initial Setup Algorithm}
\begin{algorithmic}[1]
\State \textbf{Input:} Set of NLQ $P$, golden SQL queries $G$
\State \textbf{Result:} Set of predefined examples $E$
\State $E \gets []$ \Comment{Initialize an empty set of predefined examples}
\For{each NLQ $P_i \in P$}
    \State $predicted\_sql \gets generate\_sql(P_i)$
    \If{$predicted\_sql \neq G[P_i]$}
        \State $edit\_script \gets ChangeDistiller(predicted\_sql, G[P_i])$
        \State $error \gets extract\_error(predicted\_sql)$
        \State $E.append(\{prompt: P_i, incorrect\_sql: predicted\_sql,$ 
        \State \hskip2em $golden\_sql: G[P_i], error: error, edit\_script: edit\_script\})$
    \EndIf
\EndFor
\State \textbf{return} $E$
\end{algorithmic}
\end{algorithm}

\begin{table*}[ht]
  \caption{Experiment Execution Accuracy and Fix Rate comparing two baseline approaches with the five RAG based methods proposed in this paper. The best performance is highlighted in bold.}
  \label{tab:results}
  \centering
  \begin{tabular}{lcccccc}
    \toprule
    \textbf{Method} & \multicolumn{3}{c}{\textbf{Execution Accuracy}} & \multicolumn{3}{c}{\textbf{Fix Rate}} \\
    \cmidrule(lr){2-4} \cmidrule(lr){5-7}
    & \textbf{0-shot} & \textbf{1-shot} & \textbf{3-shot} & \textbf{0-shot} & \textbf{1-shot} & \textbf{3-shot} \\
    \midrule
    Baseline without Error Correction & 72.5\% & NA & NA & 0\% & NA & NA \\
    Baseline with Simple Error Correction & 75.4\% & NA & NA & 28.9\% & NA & NA \\
    RAG using embedding of SQL Error & NA & 75.9\% & 76.2\% & NA & 34.8\% & 37.8\% \\
    RAG using embedding of NLQ & NA & 75.9\% & 76.1\% & NA & 34.5\% & 36.3\% \\
    RAG using embedding of Predicted SQL & NA & 76.2\% & 76.1\% & NA & 37.2\% & 36.6\% \\
    RAG using embedding of NLQ and Predicted SQL & NA & 75.9\% & 76.1\% & NA & 34.2\% & 36.9\% \\
    \textbf{RAG using embedding of NLQ, Predicted SQL, and SQL Error} & NA & 76.1\% & \textbf{76.4\%} & NA & 36.6\% & \textbf{39.2\%} \\
    \bottomrule
  \end{tabular}
\end{table*}

Algorithm 2 utilizes these predefined examples to correct errors in newly generated SQL queries by selecting the most relevant few-shot examples based on embedding similarities and applying the corresponding edit scripts.

\begin{algorithm}[H]
\caption{Few-Shot Selection \& Error Correction Algorithm}
\begin{algorithmic}[1]
\State \textbf{Input:} New NLQ $P_{new}$, set of predefined examples $E$
\State \textbf{Result:} Corrected SQL query
\State $predicted\_sql_{new} \gets generate\_sql(P_{new})$
\State $error_{new} \gets extract\_error(predicted\_sql_{new})$
\State $best\_match \gets find\_best\_match(P_{new}, predicted\_sql_{new},$ 
\State \hskip2em $error_{new}, E)$
\State $corrected\_sql \gets apply\_edit\_script(predicted\_sql_{new},$
\State \hskip2em $best\_match["edit\_script"])$
\State \textbf{return} $corrected\_sql$
\end{algorithmic}
\end{algorithm}

\section{Experiments}
The next subsections provide information on our dataset, LLM, and experiment setup.

\subsection{Dataset \& LLM Setup}
We utilized the open-source Gretel dataset and filtered out data where the database could not be created, the golden SQL resulted in execution errors, or the golden SQL returned empty results. This process left us with 58,193 samples in the training set and 3,425 samples in the testing set. For SQL generation, we used the mixtral-8x22b-instruct-v0.1 LLM, with DSPy for optimized prompting. Text embedding was handled using stella\_en\_1.5B\_v5, and FAISS was used for the vector store.

\subsection{Experiment Setup}
Each experiment involved generating SQL queries based on new NLQ, applying our selection process to identify the closest matching few-shots, and using the selected examples to correct the errors in the generated SQL. The results were then analyzed to determine which approach yielded the most accurate SQL corrections. To evaluate the effectiveness of our few-shot selection methodology, we conducted a series of experiments aimed at determining the best approach for selecting the closest few-shots. Specifically, we explored the following strategies:

\textbf{Few-Shot Selection Criteria:}
We focused on selecting the closest few-shot examples based on three key similarity metrics:
\begin{itemize}
\item {\texttt{NLQ Similarity}}: Measuring how closely the new NLQ aligns with those in the predefined examples.
\item{\texttt{Error Similarity}}: Assessing the similarity between the errors in the new SQL query and those in the predefined examples.
\item{\texttt{Predicted SQL Similarity}}: Comparing the generated SQL with the incorrect SQL from the predefined examples.
\end{itemize}

\textbf{Number of Few-Shots:}
We tested the impact of selecting either one or three few-shots for each approach. By comparing the performance across these variations, we aimed to identify the optimal balance between context relevance and correction accuracy.

\section{Evaluation Metrics}
This paper utilizes two evaluation metrics. The first is execution accuracy, which measures correctness by executing both the predictions and ground truths on the underlying database and comparing the results. For a prediction to be considered correct, the results must be nearly identical. The second metric is the fix rate. Since our work focuses on error correction, the fix rate measures the proportion of failed predictions due to SQL execution errors that are subsequently corrected and pass based on the standard of execution accuracy.

\section{Experiment Result}
We compare our results against a no-error correction baseline approach and a simple 0-shot error correction approach \cite{pourreza2023dinsqldecomposedincontextlearning} where the error information is supplemented to the prompt.

% \begin{table*}
%   \caption{Experiment Execution Accuracy and Fix Rate}
%   \label{tab:commands}
%   \begin{tabular}{ccl}
%     \toprule
%     Method & Execution Accuracy & Fix Rate\\
%     \midrule
%     \texttt{} & 0-shot & 1-shot \\
%     \texttt{Baseline without Error Correction}& 72.5 & NA\\
%     \bottomrule
%   \end{tabular}
% \end{table*}

The results of the experiments are presented in Table~\ref{tab:results}. We can see that error correction with RAG-few shot using embedding of NLQ, Predicted SQL and SQL Error leads to the best execution accuracy of 76.4\%, which is 3.9\% higher than the baseline model. It successfully fixes 39.2\% of the failed prediction with SQL execution errors, which is 10.3\% higher than the fix rate of Simple Error Correction. On average, 3-shot learning leads to 0.2\% higher execution accuracy and 2.0\% higher fix rate compared to 1-shot learning. While we cannot directly compare our approach’s efficacy against the Model of Code Error Correction approach \cite{chen2023texttosqlerrorcorrectionlanguage} as it was performed on a different dataset, our Fix Rate of 39.2\% is higher than the 18-21\% Fix rate observed in that study.

\section{Conclusion}
In this paper, we introduced a novel few-shot learning-based approach for error correction in SQL generation. Our method leverages embedding-based similarity measures to select contextually relevant few-shot examples, enabling more accurate SQL query generation. We achieved an execution accuracy of 76.4\%, marking a 3.9\% improvement over the baseline model without error correction. Moreover, our error correction method successfully resolved 39.2\% of failed predictions with SQL execution errors. These findings underscore the potential of embedding-based selection in enhancing the few-shot learning process, making SQL query generation more precise and reliable. Our research contributes to the ongoing development of automated SQL generation techniques, offering a robust framework for error correction that can be integrated into more advanced database interaction tools.

\section{Future Work}
Our methods only run error correction once. We believe results can be further improved if we can it multiple times. In addition, our work primarily addresses cases where SQL execution errors occur. However, only 36\% of the failed predictions are due to SQL execution errors, while the remaining 64\% fail because of mismatched results compared to the golden SQL outputs. Future work will explore reflection mechanisms even in the absence of SQL execution errors. This could involve developing an agentic pipeline to determine the most suitable embedding and leveraging an LLM as a judge to iteratively refine the SQL, assessing the predicted SQL and its results at each step to ensure they make sense. Furthermore, we believe that the retrieval of few-shot examples can be enhanced by utilizing state-of-the-art embeddings and employing a hybrid retrieval approach that combines both embedding-based and keyword-based similarity methods, such as BM25.

% The use of \BibTeX\ for the preparation and formatting of one's
% references is strongly recommended. Authors' names should be complete
% --- use full first names (``Donald E. Knuth'') not initials
% (``D. E. Knuth'') --- and the salient identifying features of a
% reference should be included: title, year, volume, number, pages,
% article DOI, etc.

% The bibliography is included in your source document with these two
% commands, placed just before the \verb|\end{document}| command:
\bibliographystyle{ACM-Reference-Format}
\bibliography{sample-base.bib}

\end{document}